%% file: main.tex
\documentclass[letterpaper, 10 pt, journal, twoside]{IEEEtran}
\usepackage{amsmath,amsfonts}
\usepackage{url}
\usepackage{graphicx}
\usepackage{cite}

\usepackage{pifont}
\newcommand{\cmark}{\ding{51}}%
\newcommand{\xmark}{\ding{55}}%

\usepackage{booktabs}
\usepackage{multirow}
\usepackage{subfigure}

\usepackage{siunitx}
\usepackage{threeparttable}
\usepackage[dvipsnames]{xcolor}

\usepackage{xspace}
\makeatletter
\DeclareRobustCommand\onedot{\futurelet\@let@token\@onedot}
\def\@onedot{\ifx\@let@token.\else.\null\fi\xspace}

\def\eg{\emph{e.g}\onedot}

\def\etc{\emph{etc}\onedot}

\def\etal{\emph{et al}\onedot}
\makeatother

\begin{document}

\title{Panoptic Vision-Language Feature Fields}

\author{Haoran Chen$^{1}$, Kenneth Blomqvist$^{1}$, Francesco Milano$^{1}$, and Roland Siegwart$^{1}$
\thanks{Manuscript received: August 31, 2023; Revised November 28, 2023; Accepted January, 1, 2024.}
\thanks{This paper was recommended for publication by Editor Markus Vincze upon evaluation of the Associate Editor and Reviewers' comments.
This work has received funding from the European Union’s Horizon 2020 research and innovation program under grant agreement No. 101017008 (Harmony).}
\thanks{$^{1}$Haoran Chen, Kenneth Blomqvist, Francesco Milano, and Roland Siegwart are with the Autonomous Systems Lab, ETH Z\"{u}rich, Switzerland
{\tt\footnotesize chenhao@student.ethz.ch}}
\thanks{Digital Object Identifier (DOI): 10.1109/LRA.2024.3354624}
}


\markboth{IEEE Robotics and Automation Letters. Preprint Version. Accepted January, 2024}%
{Chen \MakeLowercase{\textit{et al.}}: Panoptic Vision-Language Feature Fields}


\maketitle

\begin{abstract}
Recently, methods have been proposed for 3D \emph{open-vocabulary} semantic segmentation. Such methods are able to segment scenes into arbitrary classes based on text descriptions provided during runtime. In this paper, we propose to the best of our knowledge the first algorithm for \emph{open-vocabulary panoptic} segmentation in 3D scenes. Our algorithm, Panoptic Vision-Language Feature Fields (PVLFF), 
learns a semantic feature field of the scene by distilling vision-language features from a pretrained 2D model, and jointly fits an instance feature field through contrastive learning using 2D instance segments on input frames.
Despite not being trained on the target classes, our method achieves 
panoptic segmentation performance similar to
the state-of-the-art \emph{closed-set} 3D 
systems on the HyperSim, ScanNet and Replica dataset and additionally outperforms current 3D open-vocabulary systems in terms of semantic segmentation. We ablate the components of our method to demonstrate the effectiveness of our model architecture. Our code will be available at \url{https://github.com/ethz-asl/pvlff}.
\end{abstract}


\section{Introduction}
\label{sec:intro}
\input{sections/00-introduction}

\section{Related Work}
\label{sec:relatedwork}
\input{sections/01-related}

\section{Method}
\label{sec:method}
\input{sections/02-method}

\section{Experiments}
\label{sec:experiments}
\input{sections/03-experiments}

\section{Conclusion and Future Work}
\label{sec:conclusion}
\input{sections/05-conclusions}

{\small
\bibliographystyle{IEEEtran}
\bibliography{egbib}
}

\end{document}

%% file: sections/00-introduction.tex
\IEEEPARstart{A}{n} important consideration for building spatial AI applications is the representation used to model the scene. Ideally, the chosen representation can be built incrementally in real-time, model the geometry with high fidelity, and allow for flexible runtime semantic queries.

Recently, \emph{open-vocabulary} semantic scene representations based on NeRF~\cite{mildenhall2020nerf} have been proposed~\cite{kerr2023lerf, blomqvist2023neural, liu20233d} for robotics. Such systems reconstruct scene geometry implicitly from 2D views, and enable zero-shot semantic segmentation and natural language-based object detection. They achieve this by distilling features from a vision-language model into a feature field representation mapping points in the scene to vision-language vectors, which can be compared against natural language prompts. 
Such representations present great promise for augmented reality, mobile manipulation, and intelligent robotic applications, bridging physical 3D spaces with natural language representations. 

A limitation of current systems is that, while they can perform \emph{semantic} segmentation, they cannot produce a \emph{panoptic} segmentation of the scene by telling instances of objects belonging to the same class apart. A key problem in 3D instance segmentations that are built from multiple views is that instance segmentations across different views are not guaranteed to be consistent. This is further complicated by the fact that 2D segments are noisy, and might only segment a subpart of each object in individual views. 

A recent approach, Panoptic Lifting~\cite{siddiqui2023panoptic}, tackles panoptic segmentation by fusing 2D panoptic segmentations into 3D. This is achieved by mapping instance identifiers across different views through a linear assignment. A drawback of this approach is that it assumes a maximum number of instances in the scene, and requires computing the assignment at each training step, which is increasingly expensive as the assumed number of instances grows. Furthermore, the semantic predictions of Panoptic Lifting are restricted to a fixed set of semantic categories.

\begin{figure}[t]
    \centering
    \includegraphics[width=\columnwidth]{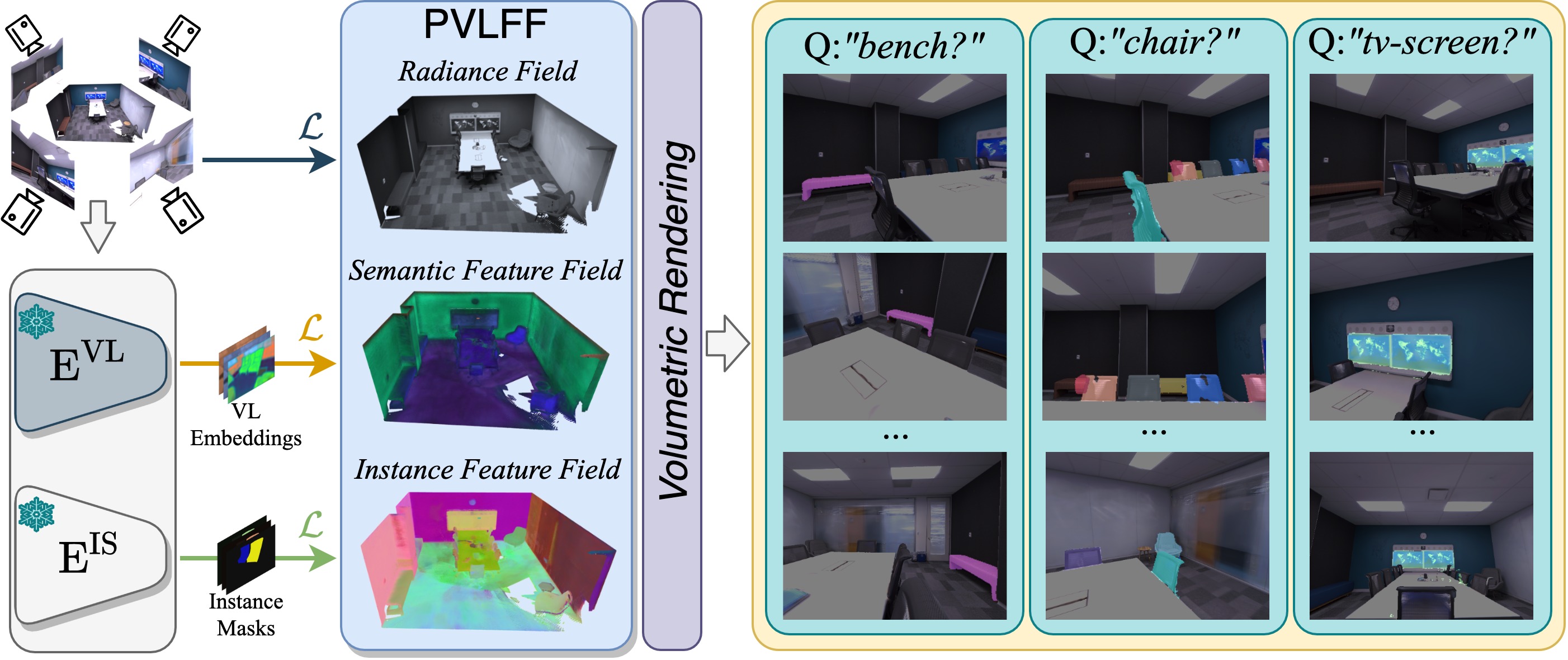}
    \caption{\textbf{Overview of PVLFF.} Given 2D posed images, PVLFF optimizes a semantic feature field by distilling vision-language embeddings from an off-the-shelf network $\mathrm{E^{VL}}$~\cite{li2022lseg}, and simultaneously trains an instance feature field through contrastive learning based on 2D instance proposals computed by $\mathrm{E^{IS}}$~\cite{kirillov2023segany}. After training through different loss functions ($\mathcal{L}$), PVLFF is able to perform panoptic segmentation under open-vocabulary prompts.}
    \vspace{-15pt}
    \label{fig:pvlff}
\end{figure}

In this work, we propose Panoptic Vision-Language Feature Fields (PVLFF), a novel pipeline for 3D open-vocabulary panoptic mapping, as shown in Fig.~\ref{fig:pvlff}. PVLFF learns a radiance field~\cite{mildenhall2020nerf}
from posed images, and simultaneously learns semantic and instance feature fields from 2D proposals computed by an off-the-shelf dense vision-language encoder $\mathrm{E^{VL}}$~\cite{li2022lseg} and a pretrained instance segmenter $\mathrm{E^{IS}}$~\cite{kirillov2023segany}. At runtime, instance features can be easily clustered using conventional clustering algorithms to produce an instance segmentation, either of the 3D pointclouds or of  2D views of the scene after volumetric rendering. The resulting instance segments can be further fused with the semantic segmentation inferred from the semantic feature field based on language prompts, producing an \emph{open-vocabulary} panoptic segmentation.

Our core insight is that existing vision-language feature field approaches~\cite{blomqvist2023neural, kerr2023lerf, liu20233d} can be extended to do panoptic segmentation by introducing a separate instance feature head, which is learned from 2D instance segments using a contrastive loss function. The contrastive loss function enables learning from inconsistent instance segments and does not require assuming a maximum instance count. 
We furthermore show that the learned instance features can be clustered hierarchically, which enables segmenting instances at different scales.
Such hierarchical representation of instances can benefit scene-understanding and high-level planning applications, such as assembling a piece of furniture, which involves a robot identifying individual components like screws, bolts, or joints.

We evaluate PVLFF on the Hypersim~\cite{roberts2021hypersim}, Replica~\cite{replica19arxiv}, and ScanNet~\cite{dai2017scannet} datasets. Our method, trained with no class supervision, achieves
satisfactory results compared to
the state-of-the-art \emph{closed-set} panoptic systems. We also show that our method outperforms zero-shot methods on both 2D and 3D semantic segmentation ($+4.6\%$ of $\mathrm{mIoU}$). We further ablate the design choices of our method.

In summary, our contributions are:
\begin{itemize}

    \item A hierarchical instance feature field that enables obtaining multi-scale 3D instance segments from 2D proposals using contrastive learning;
    \item To the best of our knowledge, the first zero-shot \emph{open-vocabulary} 3D panoptic segmentation system.

\end{itemize}


%% file: sections/01-related.tex
\textbf{Panoptic Segmentation.} The task of (2D) panoptic segmentation was first introduced by Kirillov~\etal~\cite{kirillov2019panoptic} to provide a unified vision system that would produce coherent segmentations for both \textit{stuff} -- generic amorphous regions, the focus of semantic segmentation -- and \textit{things} -- countable objects, that instance segmentation works aim to delineate. Specifically, the goal of panoptic segmentation is to assign a semantic and an instance label to each pixel in an image~\cite{kirillov2019panoptic}.
After a first wave of works~\cite{cheng2022masked, kirillov2019panoptic, cheng2020panoptic} in 2D panoptic segmentation, numerous works have explored panoptic segmentation in a 3D context. \cite{fong2022panoptic, sirohi2021efficientlps, schult2022mask3d} take 3D structures (\eg, point cloud, mesh) as input to predict 3D panoptic segmentation. \cite{grinvald2019volumetric, dahnert2021panoptic, narita2019panopticfusion} propose to simultaneously segment a 3D scene and reconstruct the geometry from 2D images. Recently, NeRF-based methods~\cite{siddiqui2023panoptic, fu2022panoptic, KunduCVPR2022PNF, bhalgat2023contrastive} have achieved state-of-the-art performance on 3D benchmarks. A concurrent NeRF-based work, Contrastive Lift~\cite{bhalgat2023contrastive}, performs panoptic segmentation by training an instance feature field from 2D proposals using contrastive learning. However, it can only classify specific semantic categories and is restricted to pre-defined classes for instance segmentation. 
Most of these methods are task-oriented and none of them are capable of panoptic segmentation in an open set.
Instead, we propose a 3D \emph{open-vocabulary} panoptic system designed for semantic segmentation of any category and object-agnostic instance segmentation. Moreover, our method can segment objects at different scales, a capability not present in previous approaches.


\textbf{Semantic Neural Fields.} Neural fields have become an established representation for 3D scene reconstruction from 2D images ever since the introduction of NeRF~\cite{mildenhall2020nerf}, which models density and radiance for any 3D position in a scene. NeRF-based systems~\cite{deng2022depth, barron2021mip, chen2022tensorf, muller2022instant} have shown impressive results in photo-realistic rendering of novel viewpoints and accurate reconstruction of the scene geometry. Exploiting its 3D-aware nature, recent works~\cite{Zhi2021semanticnerf, tschernezki2022neural} have investigated extensions of NeRF by fusing semantic information into 3D neural fields for scene-level semantic understanding. A limitation of these methods is that they rely on 2D semantic supervision, provided in the form of labels from a fixed, closed set. Our method, in contrast, performs semantic segmentation under open-vocabulary queries, while additionally acquiring instance-aware scene knowledge.


\textbf{Open-vocabulary Scene Understanding.} In the last few years, several advances in open-vocabulary perception tasks have been achieved by leveraging CLIP~\cite{radford2021learning}, a neural network that embeds visual and language information in the same feature space using contrastive learning. \cite{xu2023side, yi2023simple, liang2023open} extended CLIP to pixel-level semantic segmentation
by generating a set of class-agnostic dense masks with corresponding class embeddings and selecting during inference the mask with embedding closest to the language query embedding. \cite{li2022lseg, ghiasi2022scaling} follow a similar scheme, but focus on per-pixel embeddings, predicting semantics according to the similarity of each pixel with the language query embedding. In addition to the 2D tasks, a number of subsequent works have proposed fusing open-vocabulary information into a 3D representation. \cite{peng2023openscene, liu20213d} perform open-vocabulary semantic segmentation on pre-computed 3D data structures. \cite{blomqvist2023neural, kerr2023lerf, liu20233d}, on the other hand, distill the open-vocabulary knowledge into a 3D representation while reconstructing scene geometry concurrently. \cite{ahn2022can} proposes a robot system with language-conditioned robotic control policies to perform complex tasks under natural language instructions. In our work, we mainly focus on panoptic scene understanding under open-vocabulary language queries by fusing pre-trained vision-language knowledge with instance information derived from our instance feature field.



%% file: sections/02-method.tex
\begin{figure}[t!]
  \centering
   \includegraphics[width=\linewidth]{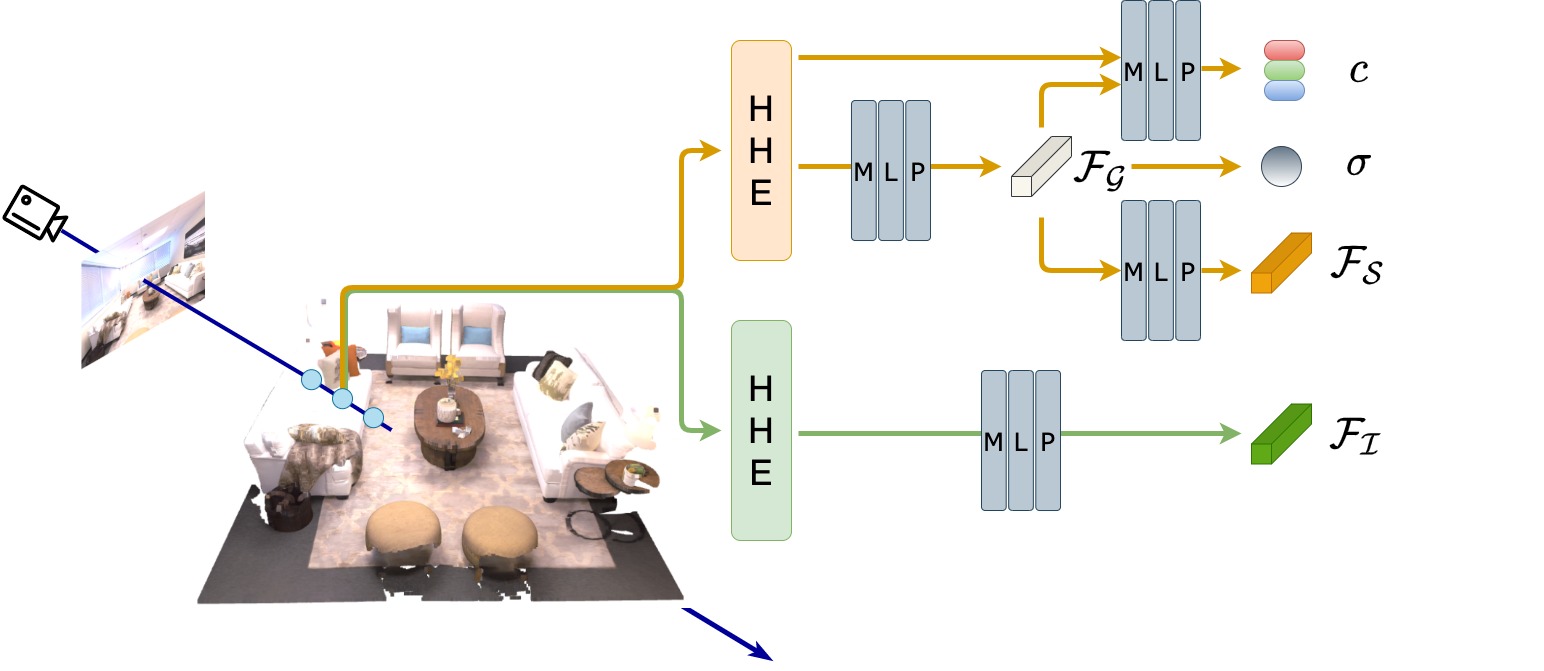}

   \caption{\textbf{Architecture of PVLFF.} Given a 3D coordinate $\mathbf{x}$ and a unit direction $\mathbf{d_r}$, PVLFF uses two sets of hybrid hash encoding (HHE)~\cite{blomqvist2022baking} to parameterize the 3D volume for panoptic scene understanding. With one HHE, we encode color $c$, density $\sigma$ and semantic feature $\mathcal{F_S}$. With the other HHE, we exclusively learn instance feature $\mathcal{F_I}$. All these scene properties are modeled by lightweight multilayer perceptrons (MLPs)}.
   \vspace{-15pt}
   \label{fig:model_overview}
\end{figure}

\begin{figure*}[t!]
  \centering
  \subfigure[Semantic feature learning.]{
  \includegraphics[width=0.46\linewidth]{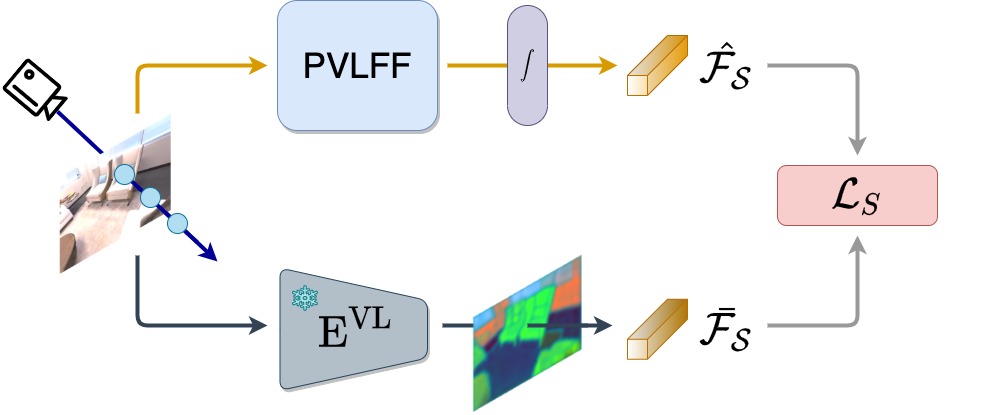}
  \label{fig:optimization-semantic}
  }
  \hfill
  \subfigure[Instance feature learning.]{
  \includegraphics[width=0.50\linewidth]{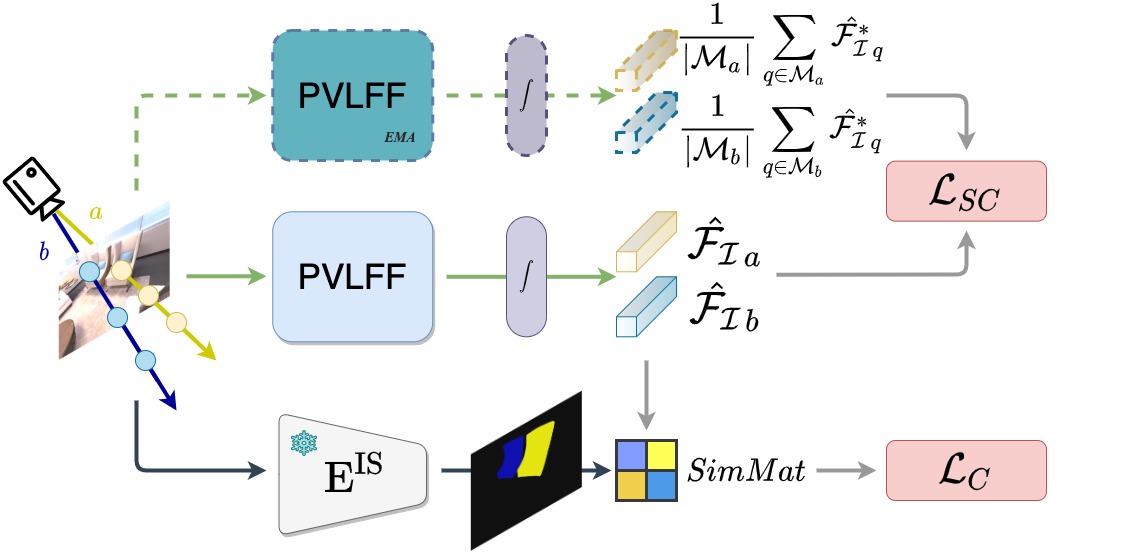}
  \label{fig:optimization-instance}
  }

  \caption{\textbf{PVLFF Optimization.} We optimize the panoptic feature fields by distilling knowledge from the off-the-shelf 2D models~\cite{li2022lseg, kirillov2023segany}. For semantic feature learning \ref{fig:optimization-semantic}, we supervise rendered semantic features with precomputed pixel-level VL embeddings. For instance feature learning \ref{fig:optimization-instance}, we pre-compute instance masks using a 2D instance segmenter~\cite{kirillov2023segany}. We then sample pixels across masks to form \emph{positive} and \emph{negative} pairs, and render corresponding instance features. We compute similarity among pairs and optimize instance features by contrastive learning. In addition, we estimate the feature center of each instance mask using instance feature field with exponential moving average (EMA) parameters and apply a $l_1$ loss between the instance features and the feature centers.}
  \vspace{-15pt}
  \label{fig:optimization}
\end{figure*}

In this Section, we present our approach, PVLFF. Given a set of posed images $\{I\}$ of a scene, our objective is to reconstruct a volumetric, implicit representation that encodes color, density, and 3D instances with associated semantics.

We use a pre-trained open-vocabulary 2D Vision-Language (VL) network~\cite{li2022lseg}, denoted as $\mathrm{E^{VL}}$,  to compute dense semantic features, and use a pre-trained 2D instance segmentation network~\cite{kirillov2023segany}, denoted as $\mathrm{E^{IS}}$, to compute instance masks for each image of the scene.
As illustrated in Fig.~\ref{fig:model_overview}, we build two branches based on Instant-NGP~\cite{muller2022instant} with hybrid hash encoding (HHE)~\cite{blomqvist2022baking}, for semantic and instance feature fields respectively.
As shown in Fig.~\ref{fig:optimization}, we train our model to learn a radiance field of the scene, while simultaneously distilling precomputed VL embeddings into semantic features and learning 3D-consistent instance features through contrastive learning using the precomputed 2D instance masks.
We then combine the semantic and instance features to perform open-vocabulary panoptic segmentation. Detailed explanations of each part are provided in the following Sections.

\subsection{Data Preprocessing}


\textbf{VL embedding extraction.} For every RGB image $I$, which is assumed to have resolution $H\times W$, we compute corresponding pixel-level embeddings from a frozen VL model $\mathrm{E^{VL}}$~\cite{li2022lseg}:
\begin{equation}
    \mathcal{\Bar{F_S}} = \mathrm{E}^\mathrm{VL} (I),
\end{equation}

\noindent
where $\mathcal{\Bar{F_S}} \in \mathbb{R}^{H\times W \times C}$ denotes the VL embeddings with $C$ channels upsampled to the same resolution as $I$, and $\mathrm{E}^\mathrm{VL}$ denotes the visual encoder in the VL model.

\textbf{Instance mask extraction.} We compute instance masks for every image $I$ using a frozen instance segmenter $\mathrm{E^{IS}}$~\cite{kirillov2023segany}:
\begin{equation}
    \left[ \mathcal{M} \right]_k = \mathrm{E^{IS}} (I),
\end{equation}

\noindent
where $\left[ \mathcal{M} \right]_k$ denotes $k$ binary instance masks generated by $\mathrm{E^{IS}}$ for image $I$. The instance segmenter produces a set of instance segments for each frame. Note that the instance segments do not have to be multi-view consistent or cover all the pixels. Furthermore, there can be multi-level instance proposals for some pixels.

\subsection{Model Structure}
Fig.~\ref{fig:model_overview} illustrates the structure of our model.
We construct a semantic and an instance feature field based on Instant-NGP~\cite{muller2022instant}. However, unlike previous methods that stack additional feature fields directly on the density and color multilayer perceptrons (MLPs) ~\cite{siddiqui2023panoptic, fu2022panoptic, KunduCVPR2022PNF} -- which we denote as \emph{scene reconstruction} branch --, we separate the panoptic feature fields by deriving semantic features $\mathcal{F_S}$ from the NeRF ``geometric" features $\mathcal{F_G}$, similar to~\cite{blomqvist2023neural}, and simultaneously optimizing instance features $\mathcal{F_I}$ from another branch. For each of these branches, we adopt a $\mathrm{HHE}$~\cite{blomqvist2022baking}, a fast encoding method that consists of a hash grid encoding~\cite{muller2022instant} and a low frequency positional encoding to learn coarse spatial information and finer scene details without over-fitting.
Thus, we formulate panoptic feature fields as follows:
\begin{align}
    \mathcal{F_S} &:= \mathcal{F_S}\left( \mathcal{F_G} \left(\mathrm{HHE}_1(\mathbf{x})\right) \right)\label{eq:fs},\\
    \mathcal{F_I} &:= \mathcal{F_I}\left( \mathrm{HHE}_2(\mathbf{x}) \right),\label{eq:fi}
\end{align}
where $\mathrm{HHE}_1$, $\mathrm{HHE}_2$ denote two sets of $\mathrm{HHE}$, $\mathcal{F_G}$ denotes geometric features, and $\mathbf{x}$ denotes a 3D point in the scene.

\subsection{Panoptic Feature Optimization}

\textbf{Feature rendering.} Given the density field $\sigma$ from Instant-NGP~\cite{muller2022instant}, we can render features from the feature field $\mathcal{F}$ along a given ray $\mathbf{r}$ using the rendering equation \cite{mildenhall2020nerf}:
\begin{equation}
    \mathcal{\hat{F}} := \mathrm{R} (\mathcal{F} | \mathbf{r}, \sigma) = \sum_{i=1}^{N} T_i \left(1-e^{-\sigma_i \delta_i}\right) \mathcal{F}\left(\mathbf{x}_i \right),
\end{equation}

\noindent
where $\mathcal{\hat{F}}$ denotes rendered features, $\sigma_i$ denotes the predicted density of the sample point $\mathbf{x}_i$, $T_i = e^{-\sum_{j=1}^{i-1} \sigma_j \delta_j}$ denotes the transmittance along ray $\mathbf{r}$, and $\delta_i$ denotes the distance between samples. In PVLFF, the semantic and the instance feature fields are defined in Eq.~\ref{eq:fs} and Eq.~\ref{eq:fi} respectively.


\textbf{Semantic feature fusion.} Similarly to~\cite{blomqvist2023neural, kerr2023lerf}, we ground VL embeddings into a semantic feature field and derive the rendered semantic feature $\mathcal{\hat{F_S}}$. An $l_1$ loss is applied to optimize the semantic feature field, as shown in Fig.~\ref{fig:optimization-semantic}:
\begin{equation}
    \mathcal{L}_S = \left\| \mathcal{\hat{F_S}} - \mathcal{\Bar{F_S}} \right\|_1 / C,
\end{equation}
where $\mathcal{\hat{F_S}}$ is the rendered semantic features, $\mathcal{\Bar{F_S}}$ is the precomputed VL embeddings, and $C$ is the feature dimension.

\textbf{Contrastive learning of instance features.} From the instance feature field, we can predict a rendered instance feature $\mathcal{\hat{F_I}}$ for every pixel in each viewpoint. With the precomputed instance masks, we apply a variant of the PointInfoNCE~\cite{xie2020pointcontrast} contrastive learning algorithm on the rendered instance features, to encourage features within the same instance to be close while pushing away features from different instances, as shown in Fig.~\ref{fig:optimization-instance}. Specifically, for each image, we sample an \emph{anchor} pixel and a \emph{positive} pixel in every precomputed instance mask. Then, for each \emph{anchor} pixel, we additionally sample a set of \emph{negative} pixels in the same image, but outside the mask of the \emph{anchor}. To reduce the effect of the high variance during training, we follow the guiding strategy used in~\cite{jiang2021guided}, by detaching the gradients of \emph{positive} and \emph{negative} pixels. Therefore, our contrastive loss $\mathcal{L}_C$ reads as:
\begin{equation}
    \mathcal{L}_C = - \frac{1}{| \mathbf{\Omega}^+ |} \sum_{(a, p) \in \mathbf{\Omega}^+} \log \frac{\exp \left(\mathcal{\hat{F_I}}_a \cdot \mathcal{\hat{F_I}}_p^d / \tau\right)}{\sum_{(a, n) \in \mathbf{\Omega}^-} \exp \left(\mathcal{\hat{F_I}}_a \cdot \mathcal{\hat{F_I}}_n^d / \tau\right)},
\end{equation}
where $\mathbf{\Omega}^+$ and $\mathbf{\Omega}^-$ denote the index set of sampled \emph{positive} and \emph{negative} pairs. $a, p, n$ denote \emph{anchor}, \emph{positive}, \emph{negative} pixels respectively, $d$ denotes the gradient detaching operation, and $\tau$ denotes the temperature parameter.

It is worth mentioning that the contrastive loss is applied on the rendered viewpoints. Despite the absence of any form of association (\eg, instance IDs) between masks that correspond to the same instance across frames, the underlying reconstructed geometry of PVLFF naturally encourages the features of an instance in different viewpoints to have high similarity upon convergence.

In order to further encourage instance features within an instance to be close to each other, we additionally adopt a ``slow-center" strategy, inspired by the concentration loss introduced in~\cite{bhalgat2023contrastive}.
In particular, after the first training epoch, we estimate the average feature of every instance mask by querying the instance feature field. In the subsequent training epochs, we recompute the average feature and perform an exponential moving average (EMA) update, penalizing through an additional $l_1$ loss deviations between every \emph{anchor} feature and its corresponding ``slow-center":
\begin{equation}
    \mathcal{L}_{SC} = \frac{1}{| \mathbf{\Omega}^+ |} \sum_{a \in \mathbf{\Omega}^+} \left\| \mathcal{\hat{F_I}}_a - \frac{1}{| \mathcal{M}_a |} \sum_{q \in \mathcal{M}_a} \mathcal{\hat{F^{\ast}_I}}_q  \right\|_1,
\end{equation}
where $\mathcal{M}_a$ denotes the instance mask from which the \emph{anchor} $a$ is sampled, and $\mathcal{\hat{F^{\ast}_I}}$ denotes the rendered instance feature with EMA parameters.

\subsection{Inference}

For semantic segmentation, we first generate the text embeddings for the prompted labels using the text encoder of the VL model~\cite{li2022lseg}, then compute the similarity between the text embeddings of individual classes and the predicted semantic features, and finally assign each predicted feature to the class with the highest similarity score. For instance segmentation, we use a clustering algorithm to directly segment instance features. In our experiments, we use HDBSCAN~\cite{McInnes2017hdbscan}. We further fuse the instance information with the semantic information by denoising the semantics inside an instance segment using majority voting.  Our method predicts both denoised semantic segmentation and panoptic segmentation.

%% file: sections/03-experiments.tex
\begin{table*}[ht!]
\centering
\caption{Quantitative evaluation of panoptic systems. We compare our method, which is designed for \emph{open-vocabulary} queries, against the state-of-the-art \emph{closed-set} panoptic systems on $8$ Replica, $12$ ScanNet and $6$ HyperSim scenes. Our method achieves comparable performance in terms of {\upshape PQ\textsuperscript{scene} } and {\upshape mIoU}. We report the denoised semantic segmentation results of our method in parentheses.}
\label{tab:panoptic}
\begin{tabular}{lcccccc}
\toprule
\multirow{2}{*}{Method} & \multicolumn{2}{c}{Replica~\cite{replica19arxiv}} & \multicolumn{2}{c}{ScanNet~\cite{dai2017scannet}} & \multicolumn{2}{c}{HyperSim~\cite{roberts2021hypersim}} \\
\cmidrule(r){2-3} \cmidrule(r){4-5} \cmidrule(r){6-7}
                        & PQ\textsuperscript{scene}            & $\mathrm{mIoU}$         & PQ\textsuperscript{scene}            & $\mathrm{mIoU}$         & PQ\textsuperscript{scene}            & $\mathrm{mIoU}$         \\ \midrule
\textit{\scriptsize closed-set} \\
DM-NeRF~\cite{wang2022dmnerf}                & 44.1          & 56.0         & 41.7          & 49.5         & 51.6          & 57.6         \\
PNF~\cite{KunduCVPR2022PNF}                    & 41.1          & 51.5         & 48.3          & 53.9         & 44.8          & 50.3         \\
PNF + GT Bounding Boxes & 52.5          & 54.8         & 54.3          & 58.7         & 47.6          & 58.7         \\
Panoptic Lifting~\cite{siddiqui2023panoptic}       & 57.9          & \textbf{67.2}         & 58.9          & 65.2         & 60.1          & 67.8         \\
Contrastive Lift~\cite{bhalgat2023contrastive}       & \textbf{59.1}          & 67.0         & \textbf{62.3}          & 65.2         & \textbf{62.3}          & \textbf{67.9}         \\ \midrule
\textit{\scriptsize open-vocabulary} \\
Ours -- PVLFF                   & 43.5              & 57.5 (59.4)             & 44.0              & \textbf{67.1} (60.8)             & 39.4              & 50.0 (52.9)             \\ \bottomrule
\end{tabular}
\vspace{-10pt}
\end{table*}

\begin{table}[t]
\centering
\caption{Semantic segmentation of zero-shot systems. We evaluate our method and compare against the baselines on $312$ ScanNet scenes. Our method outperforms other zero-shot semantic systems on {\upshape mIoU} in both 3D and 2D segmentation tasks, while achieving comparable {\upshape mAcc}. We report the denoised semantic segmentation results of our method in parentheses.}
\label{tab:semantic}
\begin{tabular}{lcc}
\toprule
\multirow{2}{*}{Method}  & \multicolumn{2}{c}{ScanNet~\cite{dai2017scannet}} \\
\cmidrule(r){2-3}
                         & $\mathrm{mIoU}$         & $\mathrm{mAcc}$         \\ \midrule
OpenScene~\cite{peng2023openscene} - LSeg~(3D)    & 54.2         & 66.6         \\
OpenScene - OpenSeg~(3D) & 47.5         & \textbf{70.7}         \\
NIVLFF~\cite{blomqvist2023neural} (3D)              & 47.4         & 55.8         \\
Ours -- PVLFF (3D)               & \textbf{58.8} (47.5)             & 70.2 (59.3)             \\ \midrule
MSeg~\cite{MSeg_2020_CVPR} (2D)                & 45.6         & 54.4         \\
NIVLFF (2D)              & 62.5         & \textbf{80.2}         \\
Ours -- PVLFF (2D)               & \textbf{67.4} (58.9)             &  75.7 (66.6)            \\ \bottomrule
\end{tabular}
\vspace{-15pt}
\end{table}

\subsection{Experimental Setup}

For the open-vocabulary semantic features, we use LSeg~\cite{li2022lseg} to generate dense pixel-level features, and therefore set the feature dimension of $\mathcal{F_S}$ to $512$, the same value as LSeg. For the instance segmenter, we experiment with SAM~\cite{kirillov2023segany}, and set the feature dimension of $\mathcal{F_I}$ to $8$. Both feature fields are modeled by a $3$-layer MLP.

All our experiments are conducted on a Nvidia RTX 3090 GPU. We train our model for \num{20000} iterations, which takes approximately $65$ minutes on average for a given scene. At inference, it takes on average $\SI{4.1}{\second}$ to render a $480 \times 320$ image and perform panoptic segmentation according to language prompts.


\subsection{Scene-Level Panoptic Segmentation}
\label{sec:panoptic_segmentation}
To the best of our knowledge, no previous method is capable of open-vocabulary panoptic segmentation for 3D scenes. Therefore, we compare our method with previous \emph{closed-set} 3D panoptic segmentation methods to demonstrate the effectiveness of our system.

\textbf{Data.} We test on three public datasets for 3D scenes: Replica~\cite{replica19arxiv}, ScanNet ~\cite{dai2017scannet}, and HyperSim ~\cite{roberts2021hypersim}. Following the same setting as~\cite{siddiqui2023panoptic}, we remap $21$ COCO~\cite{lin2014microsoft} panoptic classes ($9$ thing + $12$ stuff) into the same class set of all datasets for evaluation. To fit a scene, we use posed RGB and depth images to optimize PVLFF, while the ground-truth semantic and instance labels are only used for evaluation.

\textbf{Metrics.} Panoptic Quality (PQ) is introduced by~\cite{kirillov2019panoptic} to evaluate panoptic segmentation on the image level. To account for the consistency of segmentation across views in a scene, \cite{siddiqui2023panoptic} proposed Scene-level Panoptic Quality (PQ\textsuperscript{scene}), by merging segments that belong to the same instance identifier for a certain scene, and computing PQ on the merged segmentation for evaluation. In our experiments, we report PQ\textsuperscript{scene} of our model on the benchmarks and additionally report the mean Intersection over Union ($\mathrm{mIoU}$) of semantic segmentation.

\textbf{Results.} We compare our model against the state-of-the-art 3D panoptic systems: DM-NeRF~\cite{wang2022dmnerf}, Panoptic Neural Fields (PNF)~\cite{KunduCVPR2022PNF}, Panoptic Lifting~\cite{siddiqui2023panoptic}, and Contrastive Lift~\cite{bhalgat2023contrastive}. 
Tab.~\ref{tab:panoptic} shows the evaluation results of our model and the baselines.
While still not fully on-par with the best baselines~\cite{siddiqui2023panoptic, bhalgat2023contrastive}, our method achieves comparable performance to state-of-the-art methods~\cite{wang2022dmnerf, KunduCVPR2022PNF}. Crucially, however, while all the baselines in Tab.~\ref{tab:panoptic} are \textit{supervised, closed-set} methods, PVLFF does not use any semantic or instance labels during training.
More in detail, our method produces
similar $\mathrm{mIoU}$ scores in semantic segmentation, but
lower PQ\textsuperscript{scene} scores in panoptic segmentation than, \eg,~\cite{siddiqui2023panoptic} and~\cite{bhalgat2023contrastive}.
We observe that
this is largely due to the fact that unlike these \emph{closed-set} systems, which are trained on the specific classes evaluated, PVLFF relies on the universal segmenter SAM to precompute instance masks. Due to its \emph{object-agnostic} nature, SAM is subject to over-segmentation, which causes the predicted instance masks to not match the ground-truth ones. More in general, we note that defining the boundaries of a previously \emph{unseen} instance (as is the case in open-vocabulary segmentation) is inherently an ill-posed problem. Consider for example the ceiling in the first row of Fig.~\ref{fig:visual_results}: Without previously defining whether each tile should be considered as a part of a ``ceiling" instance or as a separate instance, it is not possible to guarantee that the detected boundaries will reflect those defined in the ground-truth data. An important observation is that despite the inherent ambiguity of the problem, as we show in Sec.~\ref{sec:exp-hierarchical_features}, PVLFF predicts continuous instance features which are hierarchically structured. This hierarchy could potentially be used to produce panoptic segmentations of different granularities.


\subsection{Open-Vocabulary Semantic Segmentation}

To demonstrate the open-vocabulary capabilities of PVLFF, we compare our model against current zero-shot systems for \emph{open-vocabulary} semantic segmentation, namely MSeg~\cite{MSeg_2020_CVPR}, OpenScene~\cite{peng2023openscene} (LSeg~\cite{li2022lseg} / OpenSeg~\cite{ghiasi2022scaling}), and NIVLFF~\cite{blomqvist2023neural}. We evaluate quantitatively on the 3D semantic segmentation benchmark of ScanNet~\cite{dai2017scannet}, computing $\mathrm{mIoU}$ and mean Accuracy ($\mathrm{mAcc}$) of $20$ ScanNet classes in the validation set for 3D and 2D segmentation respectively. For neural-field methods like NIVLFF~\cite{blomqvist2023neural} and ours, we compute the 3D point cloud segmentation by querying the semantic feature for each point and then assigning semantics according to the similarities with the label text embeddings, and we compute the 2D image segmentation by segmenting rendered semantic feature maps for each viewpoint.

\textbf{Evaluation on ScanNet.} 
For point cloud segmentation (3D), we predict semantic labels for each ground-truth point on the scene level. For image segmentation (2D), we predict image semantic segmentation for each viewpoint of every scene. Note that OpenScene is given the advantage of using ground-truth 3D point cloud, while NeRF-based methods, like NIVLFF and ours, reconstruct the scene geometry implicitly from 2D posed images. As shown in Tab.~\ref{tab:semantic}, our model outperforms state-of-the-art methods in \emph{open-vocabulary} point cloud segmentation, with the best $\mathrm{mIoU}$ and the near-best $\mathrm{mAcc}$. 
In terms of image segmentation on rendered images, PVLFF produces the best \emph{open-vocabulary} semantic segmentation results, with a large leading gap ($+4.9\%$) in $\mathrm{mIoU}$ compared to the baselines.
 
\begin{figure*}[t!]
  \centering
  \includegraphics[width=\textwidth]{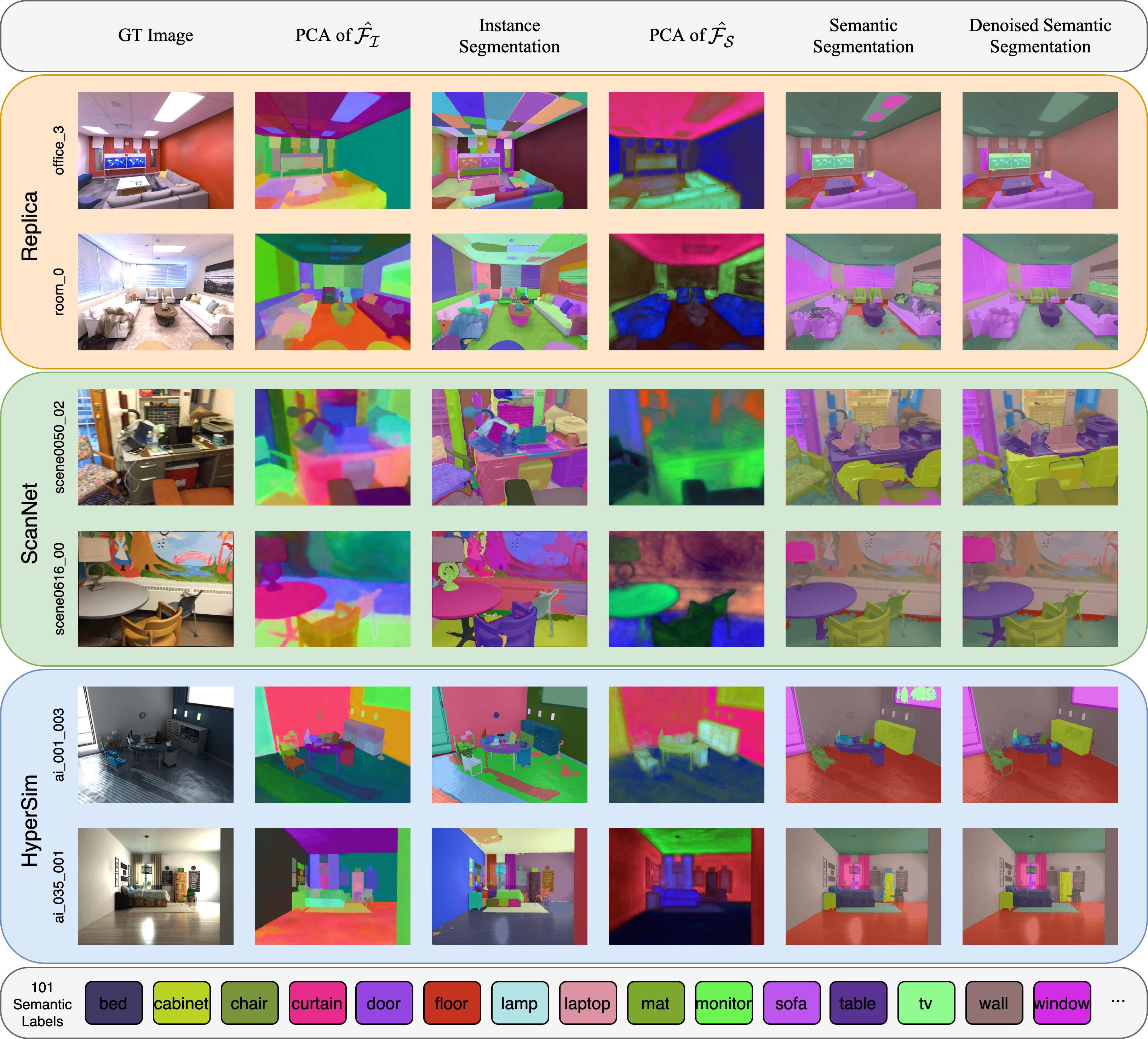}
  \caption{\textbf{PVLFF with Open-Vocabulary language queries.} We query PVLFF with $101$ Replica~\cite{replica19arxiv} semantic prompts, and visualize the instance and semantic features through PCA. We show the instance segmentation results directly from HDBSCAN~\cite{McInnes2017hdbscan} and the semantic segmentation together the denoised one.}
  \label{fig:visual_results}
  \vspace{-10pt}
\end{figure*}
 
\textbf{Qualitative Visual Results.} In Fig.~\ref{fig:visual_results} we present the visual results of PVLFF on different datasets with language prompts of $101$ Replica~\cite{replica19arxiv} classes. The instance feature field can segment scenes into multiple instances with good quality, and can even segment based on textures. For example, in $\texttt{scene0616\_00}$, PVLFF segments $\texttt{wall}$ into different parts based on the paintings on the wall. For semantic segmentation, PVLFF can predict rare categories correctly that are not recognized by the \emph{closed-set} panoptic systems (Sec. \ref{sec:panoptic_segmentation}), such as $\texttt{laptop}$, $\texttt{mat}$, $\texttt{monitor}$, \etc. However, the visual encoder of LSeg \cite{li2022lseg} is trained on ADE20K \cite{zhou2017scene}, a small closed-set dataset. Therefore, PVLFF inherently performs worse in certain categories, such as $\texttt{lamp}$ and $\texttt{door}$. In this sense, LSeg becomes the bottleneck of our system, which we would like to address in future work.

\subsection{Hierarchical Instance Features}
\label{sec:exp-hierarchical_features}

\begin{figure*}[ht!]
  \centering
  \subfigure[Hierarchical clustering of instance features.]{
  \includegraphics[width=0.38\linewidth]{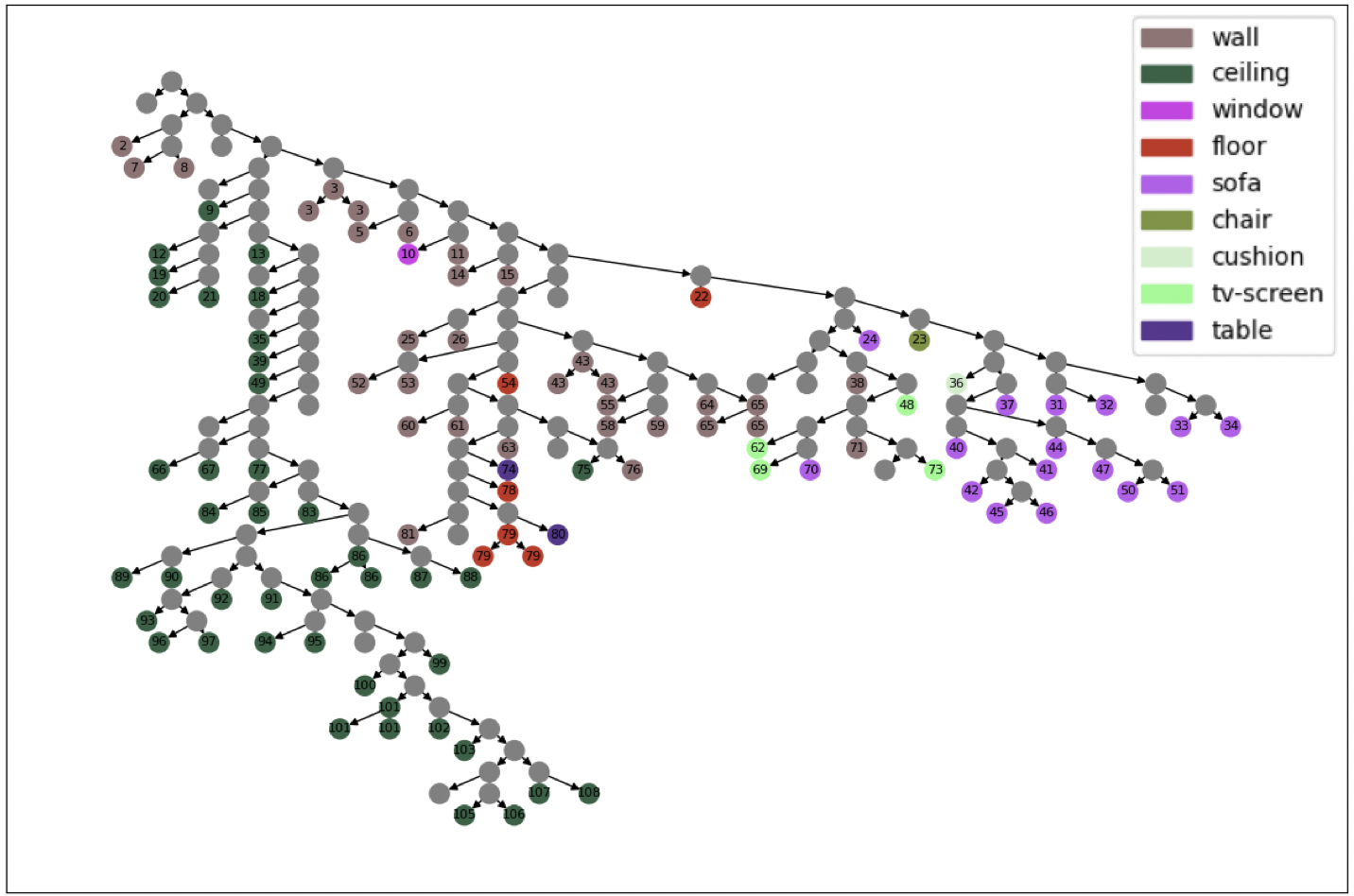}
  \label{fig:hierarchical-tree}
  }
  \hfill
  \subfigure[Hierarchical instances of ``sofa" and ``ceiling".]{
  \includegraphics[width=0.58\linewidth]{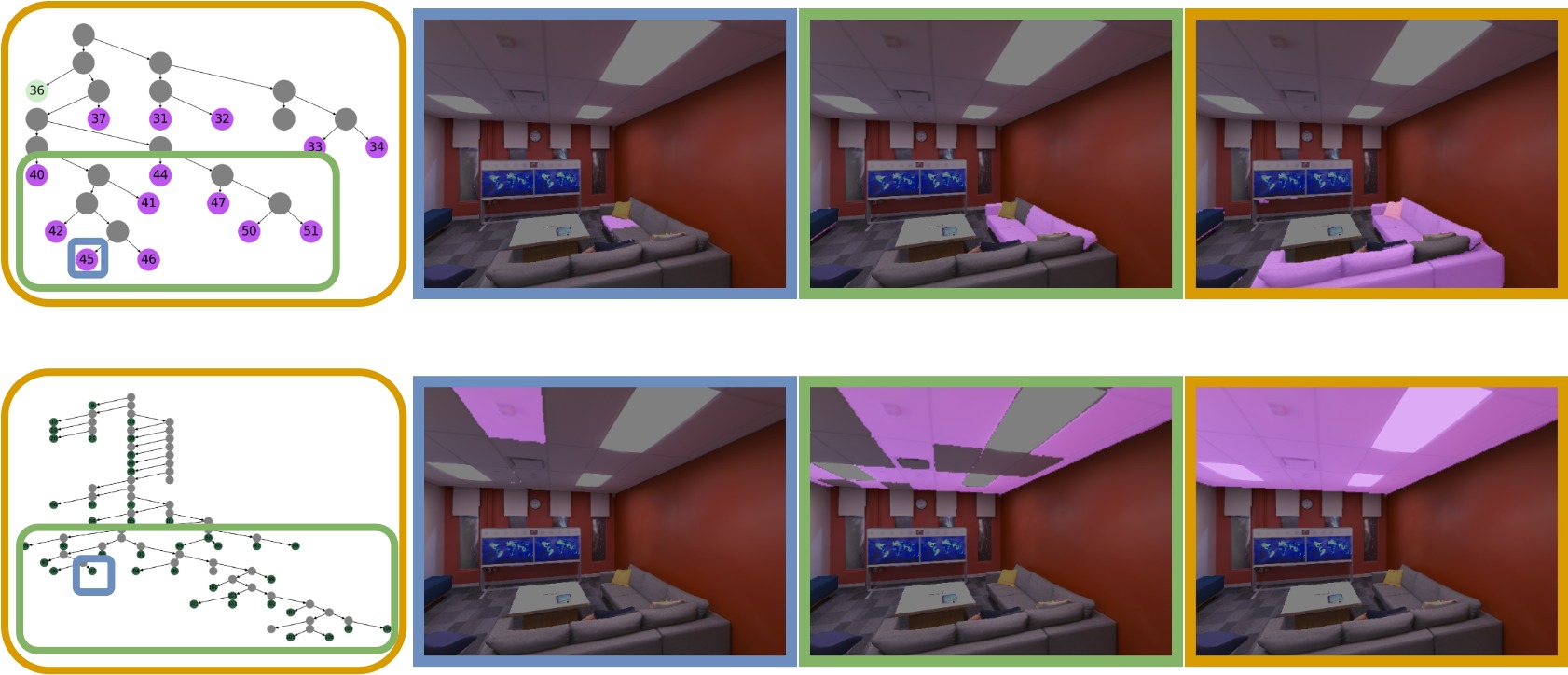}
  \label{fig:hierachical-sofaceiling}
  }
   \caption{\textbf{Hierarchical instance features of PVLFF.} We run HDBSCAN on rendered instance features and visualize the clustering results. In the clustering tree, each colored node represents a predicted instance. Since we compute instance masks using SAM, which produces multiple levels of segmentation, PVLFF over-segments instances by default. However, we can recover different levels of instance predictions through clustering and we show the multi-level predictions of ``sofa'' and ``ceiling'' from the finest to the complete segmentation, by visualizing the \emph{leaf node} \fcolorbox[rgb]{0.43,0.56,0.75}{0.43,0.56,0.75}{\textcolor[rgb]{0.43,0.56,0.75}{ff}}, the \emph{mid-part} \fcolorbox[rgb]{0.51,0.70,0.4}{0.51,0.70,0.4}{\textcolor[rgb]{0.51,0.70,0.4}{ff}} and the \emph{sub-structure} \fcolorbox[rgb]{0.84,0.61,0.0}{0.84,0.61,0.0}{\textcolor[rgb]{0.84,0.61,0.0}{ff}} in the clustering tree.
}
   \label{fig:hierarchical_instance}
   \vspace{-12pt}
\end{figure*}

Since we use SAM~\cite{kirillov2023segany} as instance segmenter for 2D instance proposals, our instance feature field is built upon the \emph{object-agnostic} masks, which can be masks of multi-level parts of an instance. Therefore, the instance features, after contrastive learning on 2D instance masks, exhibit a hierarchical structure, enabling instance segmentation at different scales.

In Fig.~\ref{fig:hierarchical_instance}, we show the hierarchical instance features of PVLFF and visualize different levels of instance predictions for $\texttt{sofa}$ and $\texttt{ceiling}$. With such hierarchical instance features, PVLFF provides the possibility to perform zero-shot panoptic segmentation on different granularities, enabling robotic applications that requires multi-level scene understanding.
Note that we use predictions at the finest level of clustering for the evaluations above. Potentially, an adaptive strategy for certain categories (\eg, $\texttt{sofa}$, $\texttt{bed}$), which determines the best level of instance segmentation from the clustering tree, would improve the evaluation results.

\subsection{Ablation}
\label{sec:exp-ablation}

\begin{table*}[ht!]
\centering
\caption{Ablation of design choices on HyperSim. We report the influence of our model design choices on instance ({\upshape mCov, mWCov}) and semantic ({\upshape mIoU, mAcc}) qualities, and ablate the effect of different semantic features. The denoised semantic segmentation results are presented in the parentheses.}
\begin{threeparttable}[b]
\centering
\label{tab:ablation}
\begin{tabular}{ccccccccc}
\toprule
\begin{tabular}[c]{@{}c@{}}Feature\\ $\mathrm{HHE}$\tnote{1}\end{tabular} &
  \begin{tabular}[c]{@{}c@{}}Feature\\ Decoupling\tnote{2}\end{tabular} &
  \begin{tabular}[c]{@{}c@{}}Instance\\ Feature\end{tabular} &
  \begin{tabular}[c]{@{}c@{}}Slow\\ Center\end{tabular} &
  \begin{tabular}[c]{@{}c@{}}Semantic \\ Feature\end{tabular} &
  $\mathrm{mCov}$ &
  $\mathrm{mWCov}$ &
  $\mathrm{mIoU}$ ($\mathrm{mIoU}$*) &
  $\mathrm{mAcc}$ ($\mathrm{mAcc}$*) \\ \midrule
\xmark & \xmark & \xmark & - & LSeg   & -    & -    & 47.7        & 58.7        \\
\xmark & \xmark & \cmark & \cmark & LSeg   & 56.6 & 48.4 & 50.9 (49.5) & 62.3 (58.9) \\
\cmark & \xmark & \cmark & \cmark & LSeg   & 64.8 & 52.3 & 50.6 (51.3) & 62.2 (60.8) \\
\cmark & \xmark & \cmark & \cmark & - & 66.8 & 52.9 & -           & -           \\
\cmark & \xmark & \cmark & \cmark & DINO   & 65.0 & 52.1 & -           & -           \\ 
\cmark & \cmark & \cmark & \xmark & LSeg   & 65.6 & \textbf{56.6} & 49.1 (47.1)   & 60.1 (57.3)  \\ \midrule
\cmark & \cmark & \cmark & \cmark & LSeg   & \textbf{68.3} & 54.1 & 50.0 (\textbf{52.9}) & 61.0 (\textbf{63.0}) \\ \bottomrule
\end{tabular}
\begin{tablenotes}
    \item [1] ``Feature $\mathrm{HHE}$" denotes using another {\upshape $\mathrm{HHE}$} for the underlying feature representation.
    \item [2] ``Feature Decoupling" denotes separating semantic and instance feature fields into two branches.
\end{tablenotes}
\end{threeparttable}
\vspace{-12pt}
\end{table*}

In Tab.~\ref{tab:ablation}, we show an ablation of different model designs by evaluating instance and semantic segmentation on HyperSim~\cite{roberts2021hypersim}. We investigate the influence of different design settings on the model performance, and report how different semantic features affect instance segmentation. For instance segmentation, we measure mean (weighted) coverage ($\mathrm{mCov}$, $\mathrm{mWCov}$)~\cite{wang2019associatively} to evaluate the instance-wise $\mathrm{IoU}$ of prediction for every ground-truth. 
For semantic segmentation, we measure $\mathrm{mIoU}$ and $\mathrm{mAcc}$ of both direct prediction from semantic field and denoised prediction.

We evaluate the baseline~\cite{blomqvist2023neural} (row 1 in Tab.~\ref{tab:ablation}) that stacks a VL feature head on the \emph{scene reconstruction} branch, and find that introducing instance features (row 2) improves the semantic performance compared to the baseline. We further show that by optimizing features on a different $\mathrm{HHE}$ branch (row 3), the model can largely increase the instance segmentation quality, while achieving similar semantic results. We also investigate the influence of different semantic features (row 4, 5) on the quality of instance features. Without storing any semantic information, the model (row 4) can use the full representational power of the feature $\mathrm{HHE}$ for instance features, while models with semantic features (row 3, 5) have slightly worse but still comparable instance segmentation qualities. Note that the evaluations are under open-vocabulary queries. Therefore, the model with DINO as $\mathrm{E^{VL}}$ (row 5), which does not support queries, does not predict semantic segmentation.
We further investigate the influence of the ``slow-center'' strategy.
We find that this strategy does not improve instance segmentation significantly, but we experimentally notice that the change induced to the rendering weights by backpropagating the instance feature losses results in more accurate rendered semantic features and increased semantic segmentation performance when using the ``slow-center" strategy (last row) compared to when not using it (row 6).
Finally, we show that decoupling the feature fields into different branches (last row) helps the underlying two sets of $\mathrm{HHE}$ to learn better semantic and instance information. Thus, our method (last row) can greatly improve the instance features (compared to row 5), and produces the best denoised semantic segmentation after fusing instance prediction.

%% file: sections/05-conclusions.tex
In this paper, we proposed PVLFF, a system for \emph{open-vocabulary} panoptic segmentation that reconstructs a scene implicitly as a neural radiance field, while simultaneously optimizing panoptic feature fields for scene understanding in an open set. We distill off-the-shelf vision-language embeddings into a semantic feature field, and train an instance feature field from \emph{object-agnostic} 2D proposals through contrastive learning. We showed that decoupling the features into two branches enhances the robustness and capacity of the neural scene representation. We validated our model design and evaluated against the  state-of-the-art semantic and panoptic segmentation methods on different datasets. 

An aspect that was partly studied in this work is the \emph{query-dependency} of instance segmentation in open-vocabulary panoptic segmentation systems. For example, the correct segmentations of individual keys on a keyboard \textit{vs.} the entire keyboard would be different. In our system, the instance segments are learned directly from the \emph{object-agnostic} 2D proposals, which do not take into account the query. As a consequence, we inherit the object instance bias in those precomputed 2D segmentation proposals. Future work might focus on developing a \emph{query-dependent} clustering algorithm for an interactive panoptic scene understanding system.